\documentclass{article}

\usepackage{microtype}
\usepackage{graphicx}
\usepackage{subfigure}
\usepackage{enumitem}
\usepackage{setspace}
\usepackage{booktabs} 

\usepackage{hyperref}

\usepackage[accepted]{icml2023}

\usepackage{amsmath}
\usepackage{amssymb}
\usepackage{mathtools}
\usepackage{amsthm}
\usepackage{bm}

\usepackage[capitalize,noabbrev]{cleveref}
\usepackage{siunitx}

\theoremstyle{plain}

\theoremstyle{definition}

\theoremstyle{remark}

\usepackage[textsize=tiny]{todonotes}

\icmltitlerunning{Beyond Intuition, a Framework for Applying GPs to Real-World Data}

\begin{document}

\twocolumn[

\icmltitle{Beyond Intuition, a Framework for Applying GPs to Real-World Data}



\icmlsetsymbol{equal}{*}

\begin{icmlauthorlist}
\icmlauthor{Kenza Tazi}{cam,bas}
\icmlauthor{Jihao Andreas Lin}{cam,tue}
\icmlauthor{Ross Viljoen}{cam}
\icmlauthor{Alex Gardner}{jpl}
\icmlauthor{ST John}{aalto}
\icmlauthor{Hong Ge}{cam}
\icmlauthor{Richard E. Turner}{cam}
\end{icmlauthorlist}

\icmlaffiliation{cam}{Department of Engineering, University of Cambridge, Cambridge, UK}
\icmlaffiliation{aalto}{Department of Computer Science, Aalto University, Espoo, Finland}
\icmlaffiliation{jpl}{Jet Propulsion Laboratory, California Institute of Technology, Pasadena, CA, USA}
\icmlaffiliation{bas}{British Antarctic Survey, Cambridge, UK}
\icmlaffiliation{tue}{Max Planck Institute for Intelligent Systems, Tübingen, Germany}

\icmlcorrespondingauthor{Kenza Tazi}{kt484@cam.ac.uk}

\icmlkeywords{Gaussian Processes, GP, kernel design, real-world data, scalability, glacier elevation change, machine learning, Bayesian Inference}

\vskip 0.3in

]

\printAffiliationsAndNotice{}




\begin{abstract}
Gaussian Processes (GPs) offer an attractive method for regression over small, structured and correlated datasets. However, their deployment is hindered by computational costs and limited guidelines on how to apply GPs beyond simple low-dimensional datasets. We propose a framework to identify the suitability of GPs to a given problem and how to set up a robust and well-specified GP model. The guidelines formalise the decisions of experienced GP practitioners, with an emphasis on kernel design and options for computational scalability. The framework is then applied to a case study of glacier elevation change yielding more accurate results at test time.
\end{abstract}

\section{Introduction}
\label{introduction}

A Gaussian Process (GP) is a probabilistic machine learning model usually used for supervised regression or classification tasks. GPs offer many advantages to modellers. First, they are suitable for small, correlated, and non-gridded datasets with missing values. Second, GPs provide a way to encode domain knowledge through prior specification and kernel design. Third and most importantly, they give principled uncertainty quantification which is critical to decision making. GPs are also more interpretable than deep learning methods. They typically have a small number of parameters that directly capture information about the data, such as the lengthscales of variation or timescales of periodicity, unlike the weights of a neural network which have no direct physical correspondence.

However, the application of GPs remains far from straightforward. In its simplest form, an exact GP is computationally expensive for large datasets, scaling cubically with the number of datapoints. Furthermore, there are comparatively few examples of GP applications suitable for non-experts. Of these examples, most are GP applications to toy datasets rather than complex real-world problems. In reality, many decisions relating to model design and implementation are developed through years of practice, reading academic literature, re-implementing experiments in code repositories, and detailed knowledge of software packages supported by small communities.

This paper aims to formalise the intuition of GP practitioners. We present guidelines to address two questions: {\it (i)} Is GP regression both computationally and methodologically applicable to a given problem? {\it (ii)} If so, how should the GP model be designed? This review is not an exhaustive list of possible GP applications but offers a starting point for using ‘application grade’ methods straightforwardly with current open-source software. For this reason, many recommendations follow a simple but principled approach to dealing with the challenges of applying GPs to real-world data. This framework can also be used as a guide when applying GP as a baseline to more complex models where simply choosing a squared exponential kernel does not showcase the performance achievable with a GP. We first define exact GP regression (\cref{sec:def}) and relate this work to previous research (\cref{sec:related}). We then present the framework (\cref{sec:framework}) and apply it to a case study of glacier height change using highly-structured satellite data (\cref{sec:study}).

\section{Gaussian Processes}
\label{sec:def}

\subsection{Definition}

Consider the set of observations comprising of input-output pairs $\{{\bm{x}_i, y_i}\}$ with $i=\{1,..., N\}$, $\bm{x}_i \in \mathbb{R}^D$ and $y_i \in \mathbb{R}$. These observations are generated by a function $f$, describing the relationship between inputs and outputs, and modulated by a noise term that accounts for the uncertainty in the observed data:
\begin{equation}
y_i = f(\bm{x}_i) + \epsilon_{i} ,
\label{eq:problem}
\end{equation}
where $\epsilon_{i}$ is assumed to be distributed normally as $\mathcal{N}(0; \sigma_n^{2})$ with standard deviation $\sigma_n$. The latent function $f$ can be modelled with a Gaussian Process (GP) \citep{rasmussen2006gaussian}. A GP is a stochastic process where any finite collection of its random variables is distributed according to a multivariate normal distribution. Generalising to infinity, a GP can be viewed as a distribution over functions. A GP is defined by a mean function $\mu(\bm{x})$ and covariance or kernel function $k(\bm{x}, \bm{x}^{\prime})$:
\begin{equation}
f(\cdot) \sim \mathcal{GP}( \mu(\bm{x}; \bm{\theta}_\mu), k(\bm{x}, \bm{x}^{\prime}; \bm{\theta}_k)) ,
\end{equation}
where both mean and kernel function typically depend on hyperparameters $\bm{\theta}_\mu$ and $\bm{\theta}_k $. 

A standard method for learning of the kernel hyperparameters is to maximise the marginal likelihood, the probability density of the observations given the hyperparameters. The marginal likelihood is computed by integrating over the values of $f$. Collecting inputs and outputs into $\bm{X} = (\bm{x}_i)_{i=1}^N$ and $\bm{Y} = (y_i)_{i=1}^N$, the logarithm of the marginal likelihood is given by
\begin{multline}
    \log (p(\bm{Y} \vert \bm{X}, \bm{\theta})) = -\frac{1}{2} (\bm{Y} - \bm{\mu})^{\top} (\bm{K}+\sigma_{n}^2 \bm{I})^{-1} (\bm{Y}-\bm{\mu}) \\
    - \frac{1}{2} \log (\vert \bm{K} + \sigma_{n}^{2} \bm{I} \vert) - \frac{n}{2} \log (2 \pi) ,
\label{eq:logmarginallikelihood}
\end{multline}
where the mean vector $\bm{\mu}$ collects $(\mu(\bm{x}_i))_{i=1}^N$ and the kernel matrix $\bm{K}$ is constructed from $k(\bm{x}_i, \bm{x}_j)$ evaluated on all pairs $i,j = 1,\dots,N$.
The hyperparameters of mean and kernel function and of the likelihood ($\sigma_n^2$) are collected in the hyperparameter vector $\bm{\theta}$.
Maximising \cref{eq:logmarginallikelihood} w.r.t $\bm{\theta}$ then gives the Maximum Likelihood Estimate for the hyperparameter values.

Assuming a Gaussian likelihood for $\bm{\epsilon}$ (see \cref{eq:problem}), the posterior predictive distribution is tractable and can be used to calculate predictions for a new output $f_*$, given a new input $\bm{x}_*$, as
\begin{equation}
p(f_* \vert \bm{Y},\bm{X},\bm{x}_*) = \mathcal{N}(f_* \vert \mu_*(\bm{x}_*), \sigma_{*}^2 (\bm{x}_* )) .
\end{equation}
Predictions are computed using the predictive mean $\mu_*$, while the uncertainty associated with these predictions is quantified through the predictive variance $\sigma_{*}^2$:
\begin{equation}
\begin{aligned}
\mu_* (\bm{x}_*)
    &= \bm{k}_{*n}^{\top} (\bm{K}+ \sigma_n^2 \bm{I})^{-1} (\bm{y} - \bm{\mu}) + \mu(\bm{x}_*) , \\
\sigma_{*}^2 (\bm{x}_*)
    &= k_{**} - \bm{k}_{*n}^{\top} (\bm{K} + \sigma_n^2 \bm{I})^{-1} \bm{k}_{n*} ,
\end{aligned}
\label{eq:predictions}
\end{equation}
where $ \bm{k}_{*n} = [k(\bm{x}_*, \bm{x}_1), \dots, k(\bm{x}_*, \bm{x}_n)]^{\top} $
and $k_{**} = k(\bm{x}_*,\bm{x}_*)$.

\subsection{Strengths}

GPs have a number of advantages that can make them a judicious choice over other supervised machine learning algorithms. A GP provides:
\begin{itemize}[topsep=0pt, itemsep=0pt]
    \item \emph{Well-calibrated uncertainty estimates.} Assuming the specified model is appropriate, a GP ‘knows when it does not know’, increasing the uncertainty away from the training distribution. This is useful for determining the likelihood of extreme events and improving decision making.
    \item \emph{Machine learning for correlated datasets.} In many cases such as geospatial problems, neighbouring observations are usually not independent and identically distributed (i.i.d.) but closely correlated to one another \citep{lalchand2022kernel, bhatt2017improved}.
    \item \emph{More interpretable machine learning.} Although not as interpretable as simpler methods such as conventional linear regression or random forests, GP covariance functions specify high-level properties of the generated functions which can be conveyed in natural language \citep{lloyd2014automatic}. Compared to deep learning methods, with thousands to trillions of learnable parameters which cannot obviously be linked with given physical features of the model predictions, GPs can be a more trustworthy alternative to the model end users.
    \item \emph{More interpretable machine learning.} Although not as interpretable as simpler methods such as conventional linear regression or random forests, GP covariance functions specify high-level properties of the generated functions which can be conveyed in natural language \citep{lloyd2014automatic}. Compared to deep learning methods, with thousands to trillions of learnable parameters which cannot obviously be linked with given physical features of the model predictions, GPs can be a more trustworthy alternative to the model end users.
    \item \emph{Data-efficient machine learning.} A GP is a non-parametric Bayesian method which provides both model expressivity while avoiding overfitting. GPs adapt to different datasets and handle data efficiently without requiring a predefined model structure. Furthermore, once a GP model is trained, new data points can be incorporated efficiently without retraining \citep{bui2017streaming}.
    \item \emph{Machine learning systems.} GP regression is unlikely to fail and can be used reliably as a subpart of a bigger machine learning system, for example in probabilistic numerics \citep{hennig2022probabilistic}, reinforcement learning \citep{engel2005reinforcement} and automated statisticians (see \cref{sec:related}).
\end{itemize}

\subsection{Limitations}

It is also important to acknowledge the limitations of GPs. They struggle with:
\begin{itemize}[topsep=0pt, itemsep=0pt]
     \item \emph{Large numbers of datapoints}. Training on datasets with $N \gtrsim 10^{4}$ becomes prohibitive \citep{deisenroth2015distributed, wang2019exact}. The computational complexity of covariance matrix inversion in \cref{eq:logmarginallikelihood} and \cref{eq:predictions} scales as $\mathcal{O}(N^3)$. The memory for storing the matrices scales as $\mathcal{O}(N^2)$.
     \item \emph{High-dimensional input spaces}. raining on datasets with $D \gtrsim 100$ becomes difficult due to the need to compute pair-wise elements of covariance function which scales as $\mathcal{O}(DN^2)$, e.g.~they are not best-suited to images \citep{van2017convolutional}. 
     \item \emph{Complex covariance functions}. In situations which require covariance functions with many parameters that must be learned from the data, the covariance function will be hard to design and the model may overfit \citep{ober2021promises}.
     \item \emph{Non-Gaussian distributions}. While Gaussian prior and likelihood function assumptions are quite common in classical scientific computing techniques, modern models, such as deep generative models, are increasingly moving towards modelling the target prior/posterior distributions using more flexible distributions parameterised by deep neural networks, like normalising flows \citep{korshunova2018bruno, casale2018gaussian}. These models avoid explicit handcrafted assumptions about the data or model and enable less biased Bayesian inference.
     \item \emph{Misspecified model}. Here, misspecification refers to our belief, or lack thereof, in the proposed model to accurately represent the underlying patterns present in the data. In this case it's hard for the model to generate accurate results. For example, an inappropriate kernel function could be chosen for the covariance matrix. Poorly specified models will not only produce a more inaccurate mean posterior distribution but also inaccurate confidence intervals and inappropriate samples \citep{sollich2001gaussian, sollich2004can, brynjarsdottir2014learning}.
\end{itemize}

\section{Related work}
\label{sec:related}

Research focused on overcoming the limitations of GPs is vast, from improving scalability \citep{liu2020gaussian}, to overcoming the model selection problem \citep{liu2020task,simpson2021kernel}. However, these methods are usually not beginner-friendly and in most cases applied to clean and well-studied benchmark datasets.
Previous work on the democratisation of GPs is centred around creating an Automatic Statistician \citep{steinruecken2019automatic} which takes in data and outputs results and a model fit in natural language. This framework builds on Automated Bayesian Covariance Discovery \citep{lloyd2014automatic} which uses the Bayesian Information Criterion to brute force the design of sensible kernel function. However, this endeavour sets aside one of the main advantages of GPs: incorporating prior knowledge. It also does not educate modellers about how to use GPs and therefore properly interpret their results.

Instead, our work follows a similar structure to other data science and machine learning guidelines with supporting code and examples to empower the deployment of GPs in the real world \citep{yu2020veridical, bell2022modeling}.

\section{Framework}
\label{sec:framework}

The following section gives an overview of the framework, illustrated in Figure \ref{fig:flowchart}, with the complete framework detailed in Appendix \ref{sec:detailed_framework}. 

As with any data science problem, the first course of action is to define the task at hand (\textbf{Step 1}). What kind of predictions would we like to make? Will the model interpolate or extrapolate the training data? What should the model output be? Are uncertainties needed? Is the output conditional on other variables? An initial exploration of the data can then be performed (\textbf{Step 2}). The structure, size and dimensions of the dataset should be identified. These first two steps should help the modeller identify the suitability of applying GPs to the task at hand  and the best way to set up the model evaluation.
\begin{figure}[ht]
    \begin{center}
        \includegraphics[width=0.9\columnwidth]{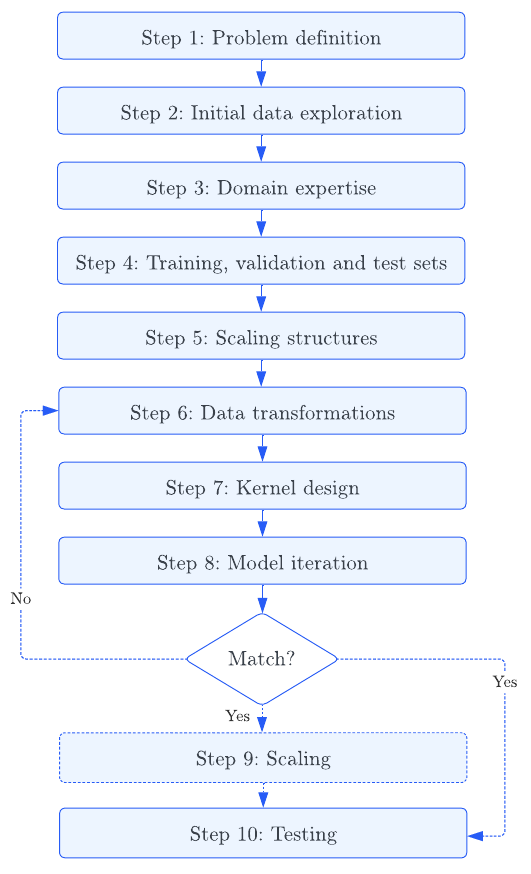}
    \end{center}
    \caption{Flowchart of the GP framework. Steps 5 and 9 can be skipped if scaling is not required.}
    \label{fig:flowchart}
\end{figure}

\textbf{Step 3} formalises the domain and prior knowledge of the modeller in a systematic way. This can help identify  early on inputs with strong predictive power, kernel structure such as periodicity, and initialisations or priors for hyperparameters. 

In \textbf{Step 4}, from the information collected thus far, the training, validation and test sets can be selected as independently as possible with respect to the problem that is trying to be addressed. More in-depth analyses can now be performed on the data.

First, if the dataset is large, we can identify scaling structures in the data in \textbf{Step 5}. We suggest three straightforward scaling cases that have robust approximation schemes and existing open-source code: over-sampled functions, timeseries data, and Kronecker covariance structure. If no scaling structure is obvious, dividing the data into manageable chunks using an unsupervised clustering method is a baseline alternative.

Second, data transformations can be ascertained and applied (\textbf{Step 6}). To improve inference, input features should be z-scored. Exact GPs also expect the posterior distribution to be Gaussian. Applying a transformation to the target data, i.e. the marginal likelihood, can help achieve this. Transforms can also help avoid unphysical predictions such as negative values and highlight the areas of the distribution we would like predict with more accurately.

Third, we can analyse the properties of that data that will inform kernel design (\textbf{Step 7}). The smoothness, covariance lengthscales, periodicity, outliers and tails, asymmetry, and stationarity should be assessed for the target variable. Once these features have been identified, the kernel can be built through composition, i.e.~adding and multiplying standard base kernels. The most important dimensions and the simplest kernel structures should be tried first. Information from previous steps should also be used to apply constraints and priors which in many cases are the determining factor in the convergence of the model to the desired results.

To validate the kernel design (\textbf{Step 8}), the kernel should also be assembled iteratively, checking the performance of the model with each new dimension or kernel parameter. For each iteration, the physical consistency of the samples, the structure of residuals,posterior predictive likelihood scores and the Root Mean Square Error (RMSE) should be checked on both the training and validation sets. The modeller should also make use of simple non-GP baselines. The iteration process should continue until the scores start to stagnate or signs of overfitting are observed. If the results of these tests are unsatisfactory, the modeller should return to Step 6 and try a new transformation, design or set of constraints. 

Once the kernel design is determined as appropriate, the model can now be scaled using the special structure found in Step 5 or by combining independent GP models using a Bayesian Committee Machine \citep{tresp2000bayesian, deisenroth2015distributed} (\textbf{Step 9}). Finally, the metrics used for validation can be reapplied to determine performance on the test set (\textbf{Step 10}).

\section{Case study}
\label{sec:study}

The framework is now applied to the case study glacier elevation over Greenland using data derived from ICESat and ICESat-2 satellites. The implementation details are provided in Appendix \ref{sec:case_details}. 

In this regression problem, we would like to estimate glacier elevation change at unsampled locations (\textbf{Step 1}). Ocean distance, topographic elevation, slope, aspect, surface glacier velocity, and spatial coordinates $x$-$y$ are used as inputs, $D=7$ (\textbf{Step 2}). The dataset is also large with $N=5\times10^5$. We compare the framework with a pre-existing application of GPs to this problem \citep{2019AGUFM.C41A..08G}.

In the original setup, the study area is split into a \SI{30}{\km} by \SI{30}{\km} grid. The $x$-$y$ coordinates, elevation and the logarithm of the velocity are used as predictors, with kernel lengthscales initialised at \SI{30}{\km}, \SI{200}{\meter}, and \SI{0.3}{\meter\per year}, respectively. A zero mean function and squared exponential (SE) kernel with Automatic Relevance Determination (ARD) are used, except along the $x$-$y$ coordinates which are constrained to have the same lengthscale and variance. Observational noise is set at \SI{0.1}{\meter\per year}.

The main application of these results is sea level change prediction, therefore the extreme values are the most important to predict accurately. The data is collected along satellite `tracks' where neighbouring points will be highly correlated. Furthermore, observations are more dense at higher latitudes (\textbf{Step 3}). We therefore allocate randomly 70\% of tracks to the training set, 10\% to the validation set, and 20\% to the test set (\textbf{Step 4}). The data exhibits oversampling which makes this case amenable to sparse variational GPs (\textbf{Step 5}). The distribution of the variables are then examined. The glacier velocity, slope, aspect and elevation change exhibit a high degree of skewness and are transformed to more Gaussian distributions where helpful. The predictive power of the transformations is assessed using a k-nearest neighbour (k-NN) baseline. All the inputs are z-scored (\textbf{Step 6}).

The properties of the transformed data are then further investigated for the kernel design. Glacier elevation change is minimal at the centre of Greenland but the rate rapidly changes towards the coastline. Elevation and ocean distance show the strongest linear correlation with elevation change. Lengthscales are visually identified for each dimension. From this analysis, a kernel with additive structure that increases function variance near the coastline would work well (\textbf{Step 7}). We then set up a simple kernel, starting with a simple squared exponential kernel for x and y, moving towards the final design with each new kernel iteration using more variables and more complex kernel structures. As the dataset is large, we also try different chunking schemes as a function of the dataset's properties. In this case, k-means clustering results in a similar performance to arbitrarily gridding the data. The samples, residuals and metrics (shown in Table 1) are checked for the training and validation sets (\textbf{Step 8}). The final kernel is chosen to be:\looseness-1
\begin{equation}
    k = k_{\text{Mat32}}(\text{lat, lon, elev}) + k_{\text{Mat32}}(\text{ocean dist})
\end{equation}
We also apply a sparse variational GP scheme using this kernel (\textbf{Step 9}). Finally, the final results are reported on the test set (\textbf{Step 10}).

\begin{table}[t]
\setlength\tabcolsep{3.4pt}
\centering
\footnotesize
\sisetup{detect-weight}
\begin{tabular}{lccccc}
\toprule
  &  {k-NN} & {SE-ARD} & {Original} & {Sparse GP} & {Frame.} \\
\midrule
RMSE & 0.07 & 0.12 & \bf 0.06 & 0.19 & \bf 0.06 \\
RMSE 5 & 0.26 & 0.17 & 0.09  & 0.75 & \bf 0.08 \\
RMSE 95 & 0.08 & 0.22 &\bf 0.05 &  0.30 & 0.07 \\
$R^2$ & \bf 0.97 & -1.24 & -0.83 & 0.73 & 0.53 \\
MLL & {N/A} &  -1.54 & -0.05 & -0.20 & \bf -2.01 \\
\bottomrule
\end{tabular}
\caption{Comparison of model performance metrics including Root Mean Squared Error (RMSE), 5th percentile RMSE (RMSE 5) and 95th percentile RMSE (RMSE 95), coefficient of determination ($R^2$) and Mean Log Loss (MLL). The posterior distribution means are quoted for the GP models.}
\label{tab:metrics}
\end{table}

\begin{figure}[ht]
    \center
    \includegraphics[width=0.9\columnwidth]{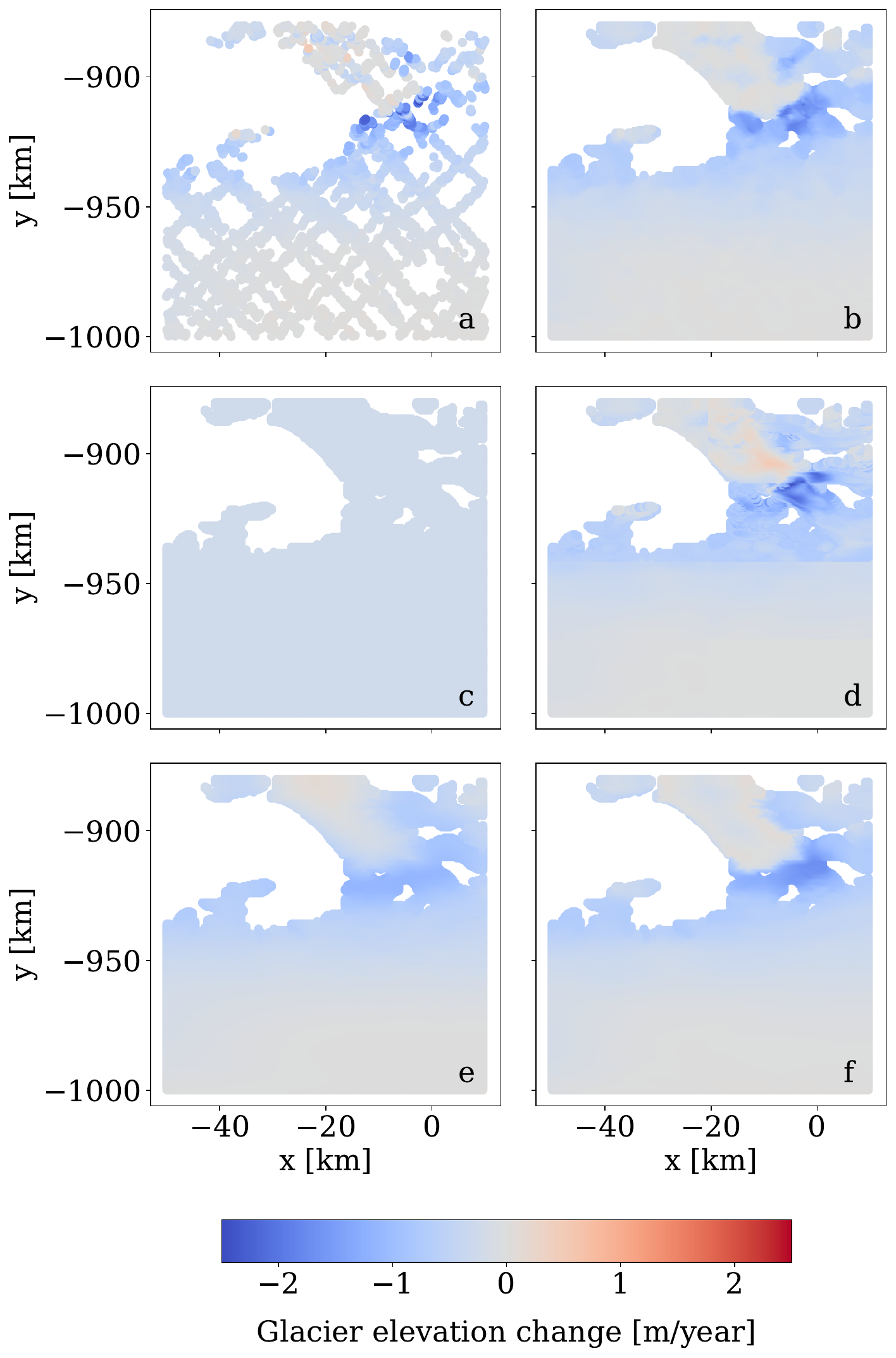}
\caption{Subset of a)the training data, with results from b) the k-NN, c) the SE-ARD GP, d) the original implementation, e) the sparse GP, and f) the framework.}
    \label{fig:results}
\end{figure}

Table \ref{tab:metrics} compares the performance of the original implementation, the framework model, the sparse GP  and two baselines: a k-NN and a GP with a SE kernel with ARD (SE-ARD). The baselines use untransformed x and y inputs only. While the k-NN model significantly outperforms the other models with respect to $R^2$, it struggles to predict the extreme negative values. The framework model and original implementation have similar RMSE scores. However, the framework model captures the variation of the data more accurately with a better $R^2$ and better constrained uncertainty estimates (low MLL). The sparse GP captures the general distribution of the model with a higher R\textsuperscript{2} but misses the extreme values we are looking to predict accurately. Figure \ref{fig:results} visually compares the model outputs for a subsection of the data. 

\section{Conclusion}

In this paper, we presented a framework to apply GP regression robustly to a wide variety of datasets while making the most of GP's strengths and working around its limitations. This formalisation of the modeller's decision process leads to improvements when applied to the case study and highlights the importance of model design when using GPs as baselines. Further work will include extending the workflow for multi-output regression and classification problems, a more in-depth analysis of GP suitability, `research grade' models and approximations, and available open-source software packages.


\section*{Software and Data}
The code, data and interactive notebooks are available at: \url{https://github.com/kenzaxtazi/icml23-gpframe.git}.


\section*{Acknowledgements}
This work was supported by the UK Engineering and Physical Sciences Research Council [grant number: 2270379] and the University of Cambridge Harding Distinguished Postgraduate Scholars Programme. The authors thank Carl Rasmussen, Will Tebbutt and Damon Wischik for their suggestions and insightful conversations. We also thank all the participants of \textit{Cambridge Stochastic Processes Workshops}, including Omer Nivron who led the first event. We are grateful to Isaac Reid for proofreading.



\newpage
\bibliography{main}
\bibliographystyle{icml2023}

\newpage
\appendix
\onecolumn
\section{Detailed framework}
\label{sec:detailed_framework}

\subsection{Step 1: Problem definition}
\label{subsec:problem}
As in any data science problem, the first step is to define the task at hand. What kind of predictions would we like to make? Will the model interpolate or extrapolate the training data? What should the model output be (e.g. a report, one model, an ensemble of models, a forecast, a de-noised timeseries)? Is the output conditional on other variables? For example, the time for a PhD to travel to the Engineering Department will depend on their start location, their mode of transport, the time of day and the time until their thesis submission. In particular, it is important to distinguish between different types of regression. Typically, problems are defined as in Section \ref{sec:def} but they could also be posed as an auto-regressive task:
\begin{equation}
    y_t = f(y_{t-1})+ \epsilon_t
\end{equation}
where $y$ at the variable $t-1$ is used to predict the observation of $y$ at the next $t$. This setup is common for emulation problems where the modeller is interested in replicating the behaviour of a system that is expensive or difficult to study in real life. The type of task which the modeller solves will determine how they design the model(s). For example, forecasting a single time-series along time in the future would mean we want to concentrate on incorporating long term dependencies present in the data. The task also shapes the metrics that should be used at validation and test time: do we care about modelling uncertainties, the mean or the extremes?

\subsection{Step 2: Initial data exploration}
\label{subsec:initial}
The next step is to perform an initial exploration of the data in order to understand whether it is suitable for GP regression. One should consider:
\begin{itemize}[topsep=0pt]
\setlength\itemsep{0.1em}
\item Number of data points $N$. For 
$N > 10^4-10^5$, exact GP computation becomes prohibitively expensive \citep{deisenroth2015distributed, wang2019exact}. $N < 100$ may be too small especially in the case of a complex kernel with many hyperparameters which can lead to overfitting. Smaller $N$ can be mitigated using Markov Chain Monte Carlo (MCMC) estimation \citep{mcneish2016using}.
\item Number of input dimensions $D$. GPs are not immune to the curse of dimensionality. For $D > 10$, it is hard for the modeller to form a clear image of the problem. It can therefore also be difficult to design an appropriate kernel. The number of parameters used to define the kernel will also increase, meaning it will be easier to overfit. Furthermore, if $D$ is very large $D>100$, constructing the covariance matrix, which scales as $\mathcal{O}(N^2D)$, can become a computational bottleneck.

\item Output requirements. Are probability distributions needed for this task? If the modeller simply requires the mean output an alternative method may be more appropriate.
\end{itemize}
Further considerations:
\begin{itemize}[topsep=0pt]
\setlength\itemsep{0.1em}
    \item  We propose a range of values for the upper limit of $N$. This is because the limit will depend on how the model is used. Higher $N$ can be used for one-off modelling rather than learning hyperparameters through repeated likelihood evaluation.
    \item Above these limits, it is worth considering the use of a GP as a wrapper for a deep learning model \citep{sun2022graph}. In this case, a deep learning model can be trained on a large subset of the data. The inputs to the GP can be the output from the deep learning model or their residuals. The GP could also be used to refine the predictions for specific locations in the input feature space using held-out data. This will then yield uncertainties which can be used for uncertainty quantification, active learning, etc.. 
    \footnote{If the models are not trained jointly, the procedure will result in overfitting and the residuals will go to zero with the GP collapsing to 0 uncertainty. If the GP is fit on the training data (and not the just the held-out data) using a neural network will be more likely to result in overfitting since the model might fit the data perfectly even before applying the GP.}
    \item It may be acceptable to work in the top end of the dimension range if only a few dimensions are doing most of the predictive work. Furthermore, it is also possible to select or generate a set of lower dimensional features to feed into the model using decision trees such as Random Forests or dimensionality reduction methods such as Principal Component Analysis \citep{binois2022survey}.
    \item For large datasets (see Step \hyperref[subsec:scalingstruct] {5}), two approaches are possible. In the first case, the data can be divided into chunks, independent GPs or `GP experts' applied are then to each part and predictions are made using a (robust) Bayesian Committee Machine \citep{tresp2000bayesian, deisenroth2015distributed}. In the second case and conditional on specific structure, scaling GP methods can be also applied.
\end{itemize}

\subsection{Step 3: Domain expertise}
\label{subsec:domain}
In this step, the modeller maps out the information that is known about the dataset, prior to a more in-depth analysis. Writing out this information in detail will be key in designing an appropriate kernel with priors and constraints in Step \hyperref[subsec:design]{7}. For example, a positive constraint is necessary when modelling rainfall.

\subsection{Step 4: Training, validation and test set definition}
\label{step:sets}

A held out dataset allows the modeller to check whether they are under or overfitting, and to give some indication of the performance they would get on `real' unseen data. The separation of the data should reflect the goals and the performance they want to measure in Step \hyperref[subsec:problem]{1}, and information from Step \hyperref[subsec:domain]{3}. Ideally the dataset should be separated into three groups: a training set, a validation set, and a test set which will not be iterated over. In many cases, the data is used in real-world is not i.i.d., it is dependent and  correlated, and come from heterogeneous sources and samplings regimes. If the effective number of data points (the number of independent samples that would be needed to produce the same information content as the given sample) is fairly small, then a cross-validation scheme is strongly recommended to assess the stability and predictability of the model \citep{yu2020veridical}.

\subsection{Step 5: Scaling structures}
\label{subsec:scalingstruct}

When working with a large dataset, the modeller should also analyse the training data to find structure that may lead to the application of scaling methods. Three cases and how to identify them are discussed below. 

\begin{itemize}[topsep=0pt]
\setlength\itemsep{0.1em}
	\item \textbf{Case 1}: Oversampled functions. An oversampled function is sampled more frequently than is required to capture its underlying structure and variation. In the case of a periodic function this would be more than the Nyquist frequency. In this situation, the modeller can use sparse GP regression with variational inference of inducing points \citep{titsias2009variational}. Many GP libraries, such \texttt{GPflow} \citep{GPflow2020multioutput} or \texttt{GPyTorch} \cite{gardner2018gpytorch}, offer built-in functions to perform this approximation.
	\item \textbf{Case 2}: Timeseries data. For timeseries data, the GP can be mapped to a Stochastic Differential Equation (SDE) \citep{hartikainen2010kalman}. This approximation has a linear cost in the number of time points and works well in many situations. However the mapping between the covariance matrix and the SDE can be expensive, sometimes more expensive than solving SDE itself. \texttt{Temporal GPs}, a package Julia, provides a framework to apply this method.
	\item \textbf{Case 3}: Kronecker product structure. For such dataset with the following kernel structure, such as gridded data:
    \begin{equation*}
        \text{Cov}\biggl(\begin{bmatrix} x_1 \\ x_2  \end{bmatrix} \begin{bmatrix} x_1^{\prime} \\ x_2^{\prime}\end{bmatrix} \biggl)=\text{Cov}(x_1,x_1^{\prime} )\otimes \text{Cov}(x_2,x_2^{\prime}),
    \end{equation*}
    the Kronecker identity can be used to invert the matrices in a piecewise fashion \citep{saatcci2012scalable}. This trick is used in the Structured Kernel interpolation (SKI) \citep{wilson2015kernel}. Both SKI and SKIP are implementable in \texttt{GPyTorch}.
 
\end{itemize}

If no structure is apparent, a baseline that is hard to beat is to the divide the data into smaller datasets, as previously mentioned. This can be done naively by separating data chronologically or into tiles. However, it can be useful to make use of clustering algorithms, such as k-means or k-nearest neighbours, to group features that exhibit similar properties.

\subsection{Step 6: Data transformations}
\label{subsec:transformations}
To improve inference, input features should generally be z-scored, i.e., subtracting the mean and dividing by the standard deviation. This means that most values should lie between -1 and 1. Exact GPs also expect the posterior distribution to be Gaussian. Applying a transformation to the target data, i.e. the marginal likelihood, can help achieve this. Transforms also can help avoid unphysical predictions such as negative values and highlight the areas of the distribution we would like to predict more accurately. Some common transformations are logarithm and power functions such as the Box-Cox transformation \citep{box1964analysis}. These transformation will constrain values to be non-zero but also reduce the weighting of the extreme values. Transformations also have a significant effect on the confidence interval of the model. The modeller should check the training data residuals and samples iterating over the chosen transform if necessary. It can be helpful to check if the transformation improves baselines such as linear regression or k-nearest neighbour models.

\subsection{Step 7: Kernel design}
\label{subsec:design}

In this step, the modeller explores the dataset in order to design the kernel function. To compose a kernel, the following characteristics of the target variable should be considered: smoothness, lengthscales, periodicity, outliers and tails, asymmetry, and stationarity. Tools such as first and second order statistics, scatter plots (with low dimensional projections), covariance and correlation matrices, autocorrelation plots, power density spectrum, and k-Nearest Neighbour analysis can be used to find these properties. The modeller should exercise their common sense when applying these tests keeping in mind the task they are trying to solve, as defined in Step \hyperref[subsec:problem]{1}. They should also view this exercise as hypothesis testing and ascertain if the characteristics of the dataset match the domain knowledge, outlined in Step \hyperref[subsec:domain]{3}, and if not, why.

This step is also important for inference time. The covariance is inverted analytically and, in most cases, Cholesky decomposition is used. However, this method requires the covariance matrix is Hermitian positive definite. This means the decomposition will fail if the one of the input dimensions is linearly related to another or a combination of other inputs. In practice, many GP models are set to have a zero mean function $\mu(\bm{x})$. However, the stability of inference can also be improved by modelling the most obvious features of the dataset through $\mu(\bm{x})$ the rather than $k(\bm{x}-\bm{x}^{\prime})$.

Now that we have a clear understanding of the data. We can design the kernel function through `kernel composition'. The kernel can be built by combining standard operators and base kernel functions \citep{duvenaud2014construction}. Adding is equivalent to applying the logical operator ‘OR’, i.e. changes in the amplitude can be explained by either term in the sum. For example, the resulting kernel will have high value if either of the two base kernels have a high value. Multiplying is similar to an ‘AND’ operation, i.e.  changes in the amplitude are explained by both term in the multiplication. For example, the kernel will have high value if both base kernels have a high value. It's worth noting that multiplying kernels can result in a lower-dimensional representation of the data, simplify eigenvalue decomposition, result in a sparse matrix, and allow for pre-computation, thus making this procedure a computationally efficient way to combine with kernels. The design of the kernel will first and foremost affect the smoothness of the samples. The modeller can enforce this by choosing the smoothness of the mean and covariance functions. This is in particular useful for modeling data with underlying structures, such as financial time series data or image data.

If the model does not initially converge close the lengthscales values determined previously. They can be initialised manually. If this still does not change where the model converges, strong guidance can be applied through constraints and priors. In real-world cases, including such specifications are not uncommon to achieve the desired results. These include:
\begin{itemize}[topsep=0pt]
\setlength\itemsep{0.1em}
    \item Boundary constraints. The modeller enforce boundary constraints on the model parameters, such as bounds on the mean function parameters or variance and lengthscale of the kernel function. This can be useful when modeling physical processes that have known limits or constraints. In practice, using constraints may be difficult. If they do not match the observations, the model can break down. It is usually better to have small and positive priors (see next point). 
    \item Bayesian priors. The model incorporate prior knowledge about the parameters of the model using priors distributions. For example, a Gaussian prior can be placed on the parameters of the covariance function to encourage the model to converge to a particular solution.
    \item Constraints on the covariance function. The modeller can enforce constraints on the covariance function, such as stationarity or monotonicity. These constraints can help to ensure that the model has physically meaningful properties.
\end{itemize}

Note that kernels with many hyperparameters will be more likely to overfit the data. Furthermore, recent literature suggests that even when a large number of terms are used, only a few parameters drive the outputs of the model after inference \citep{lu2022additive}. Regularisation techniques can be applied to the kernel function to limit overfitting \citep{cawley2007preventing} but are usually quite involved to put in place. 

\subsection{Step 8: Model iteration}
\label{subsec:iter}

The modeller should start by setting up some simple non-GP baseline such as k-NN or linear regression. They should then build the kernel iteratively, trying the most important dimensions first and check the physical consistency of the samples, the structure of residuals, the posterior predictive likelihood scores such as the marginal log likelihood, the mean log loss and Bayesian Information Criterion. The bias, Root Mean Square Error (RMSE) and Mean Absolute Error should also be evaluated. The checks should be repeated at for each iteration on both the training and validation sets. The modeller should keep iterating, until the scores start to stagnate or signs of overfitting are observed. If they are using using the GP for interpolation, this can simply be done by comparing the metrics such as RMSE and log-likelihood for the training and validation sets. However if extrapolation is the goal, it is normal for the model to perform worse away from the training distribution. In this case, it is useful to look at the GP samples. Are they sensible for both the training and validation sets? 

GP samples can also be used as `synthetic data' to check the validity of the model and training procedure. The synthetic data is generated from the model samples. This fake data is similar to real data, but where ground truth parameter values are known. The samples can be fit using another GP. If the specified model is doing a good job, the modeller should be able to recover the original covariance matrix of the training data. If the results of these tests are unsatisfactory, return to Step \hyperref[subsec:transformations]{6} and try a new transformation design or set of constraints.

If after performing Steps 6 to 8, there is no significant improvement, more involved ‘research grade’ GP methods such as a Deep Gaussian Process \citep{damianou2013deep} or a Gaussian Process Latent Variable Model \citep{pmlr-v9-titsias10a} may be more suitable for the problem.

\subsection{Step 9: Scaling}
\label{subsec:scaling}
If the modeller is using a large dataset, they can now apply their findings from Step \hyperref[subsec:scalingstruct]{5}. In particular, they will either apply one of the previously discussed scaling methods or apply independent GPs to chunked data to make predictions using a Bayesian Committee Machine. This can be implemented using the `Guepard' library \citep{guepard}.

\subsection{Step 10: Testing}
\label{subsec:testing}
Finally check the model on the test data using the same metrics as outlined in Step \hyperref[subsec:iter]{8}. These are the values that should be quoted as the results.

\section{Case study details}
\label{sec:case_details}

\begin{figure}[ht]
    \center
    \includegraphics[scale=0.4]{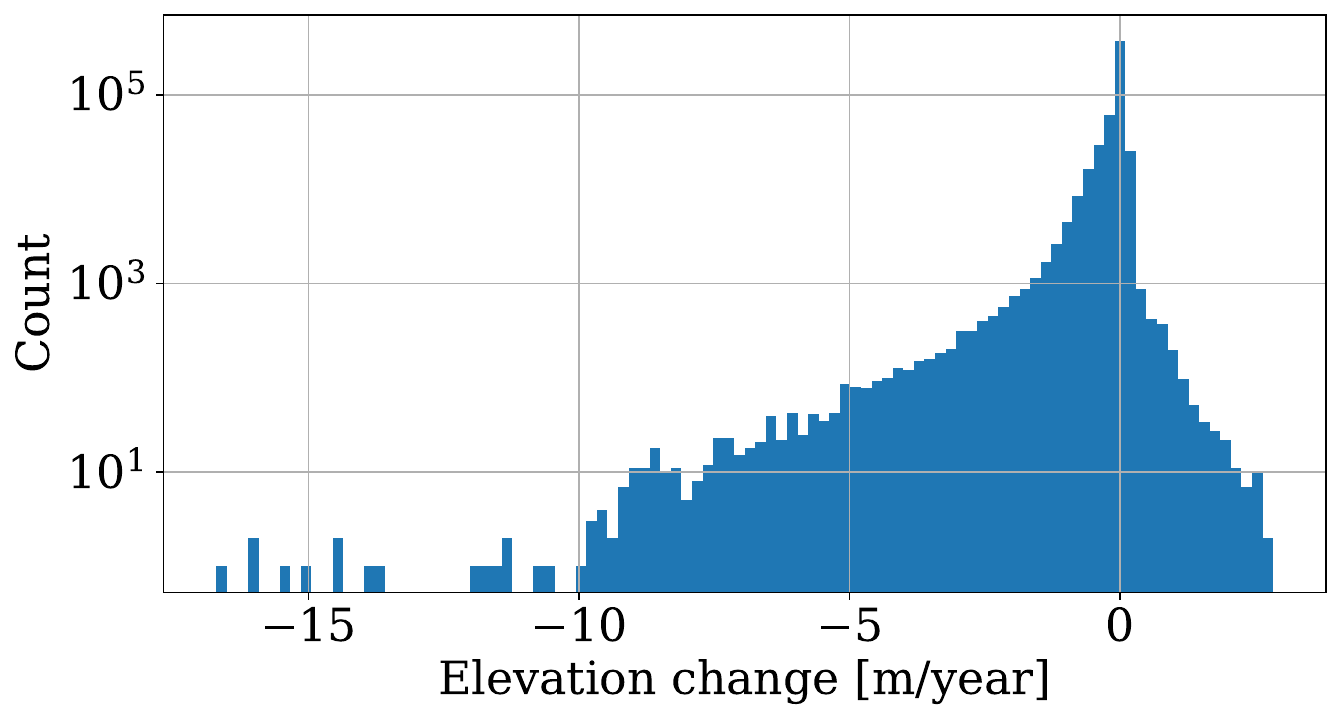} 
    \caption{Empirical probability distribution of target variable}. 
    \label{fig:pdf}
\end{figure}

The following section describes implementation details of the case study. The case study code is in implemented in \texttt{GPyTorch} with all the code executable from \texttt{Jupyter} notebooks. The most distinctive feature of the glacier elevation dataset is the distribution of the target variable shown in Figure \ref{fig:pdf}. The distribution profile change between the negative and positive elevation change. Any simple transformation struggles to make the distribution more tractable to a GP. After applying Box-Cox, and variations of logarithm and exponential transformations, the normalised raw data was still found to perform the best. 

The grid size for the independent GPs is set to approximately 130 km\textsuperscript{2}. The framework model is also initialised with the lengthscales found in Step 7. These did not help the model converge better. Priors were also applied to see if enforcing a soft constraint to keep the $x$-$y$ lengthscales more similar during optimisation helped as they tended to differ significantly for some of the grid tiles.

 Two baselines for $x$-$y$ inputs were chosen. The k-NN baseline used 10 nearest neighbours with distance weighting using the \texttt{sci-kit learn} package \citep{scikit-learn}. The k value is closely related to that of GP lengthscales \citep{vecchia1988estimation} suggesting that a `research grade' application of this dataset could include a non-stationary kernel with respect to elevation or ocean distance or the implementation of variational nearest-neigbors GPs \citep{wu2022variational}. The SE-ARD kernel was chosen as the second baseline as it is often the choice baseline for probabilistic modelling \citep{lalchand2022kernel, markou2021efficient}. This model is trained using the same parameters. 
 
 We also included a sparse GP implementation over the whole of Greenland using 1000 inducing points. Increasing the inducing beyond this point became significantly more computationally expensive, in particular, when compared to chunking with exact GP regression. The GP models were trained using conjugate gradient rather than Cholesky decoposition, for 150 epochs using ADAM with a learning rate of 0.01. The sparse GP used minibatch size of 1,024.

\end{document}